\pdfoutput=1

\documentclass[11pt]{article}

\usepackage{acl}

\usepackage{times}
\usepackage{latexsym}

\usepackage[T1]{fontenc}

\usepackage[utf8]{inputenc}

\usepackage{microtype}
\usepackage{graphicx}
\usepackage{amsmath,amssymb,amsfonts}
\usepackage{acronym}
\acrodef{nlp}[NLP]{Natural Language Processing}
\acrodef{lm}[LM]{Language Model}
\acrodefplural{lm}[LMs]{Language Models}
\acrodef{tods}[TODS]{Task-Oriented Dialog Systems}
\acrodef{ctl}[CTL]{Curricular Transfer Learning}
\acrodefplural{nn}[NNs]{Neural Networks}
\acrodef{ner}[NER]{Named Entity Recognition}
\acrodef{ai}[AI]{Artificial Intelligence}
\acrodef{rl}[RL]{Reinforcement Learning}
\acrodef{ml}[ML]{Machine Learning}
\newenvironment{plaintext}{
\ttfamily\hyphenchar\font=`\-
\spaceskip=.5em plus .5em
\xspaceskip=.5em\vspace{0.5em} \small}{\vspace{0.5em}\par}

\title{Curricular Transfer Learning for Sentence Encoded Tasks}

\author{Jader Martins Camboim de Sá\textsuperscript{1,2,*},
Matheus Ferraroni Sanches\textsuperscript{2}, \\ {\bf Rafael Roque de Souza\textsuperscript{2},
Júlio Cesar dos Reis\textsuperscript{2}, Leandro Aparecido Villas\textsuperscript{2}}\\
\textsuperscript{1} Luxembourg Institute of Science and Technology (LIST) \\
\textsuperscript{2} Institute of Computing - University of Campinas (Unicamp)\\
  \textsuperscript{*}\texttt{first.second@list.lu}
  }

\begin{document}
\maketitle
\begin{abstract}
Fine-tuning language models in a downstream task is the standard approach for many state-of-the-art methodologies in the field of NLP. However, when the distribution between the source task and target task drifts, \textit{e.g.}, conversational environments, these gains tend to be diminished. This article proposes a sequence of pre-training steps (a curriculum) guided by ``data hacking'' and grammar analysis that allows further gradual adaptation between pre-training distributions. In our experiments, we acquire a considerable improvement from our method compared to other known pre-training approaches for the MultiWoZ task.
\end{abstract}

\section{Introduction}

Traditional \ac{ml} technology is influenced by the assumption that a difference in data distribution between training and real-world data can result in a degradation of the predictive learner  \cite{shimodaira2000improving}. To overcome the effects of data divergence, studies have developed several algorithms that have some regularizing hyper-parameter to mitigate the variance in the hypothesis space and learn computable functions that have great generality to out-of-distribution data \cite{sarker2021machine}.

These traditional methods have been successfully applied in many practical applications, but they still present limitations for specific real-world scenarios where we have complex data like images or text \cite{bengio2007scaling}. Deep learning overcame a few of these limitations, acquiring near-human performance in various tasks. Whereas traditional \ac{ml} has an explicit bias to control over-fitting, deep models need to acquire its bias from data, which generally could be expensive or even unattainable in some scenarios.

Transfer learning was introduced to achieve high performance in low-resource domains by artificially increasing available data from a different domain to overcome these limitations \cite{ruder2019neural}. This technique consists in pre-training a model with high variance in domains with several available examples, so those learning models can acquire suitable biases that generalize for other tasks, including tasks where little to no examples are available \cite{Howard2018UniversalLM}.

In \ac{nlp}, the standard approach for transfer learning is language modeling, which consists in pre-training a model to denoise a sentence, and then fine-tuning it to the target task where the model is applied \cite{devlin2018bert}. This self-supervised sequential transfer learning enables the model to learn high-level linguistic features and statistical properties of language that help the model to generalize for many downstream tasks. Exploiting these capabilities, recent studies encode tasks as a pure sequence-to-sequence problem \cite{raffel2020exploring} to directly solve it with auto-regressive language modeling task \cite{Brown2020LanguageMA}.

\begin{figure}[ht]
    \centering
    \includegraphics[width=\columnwidth]{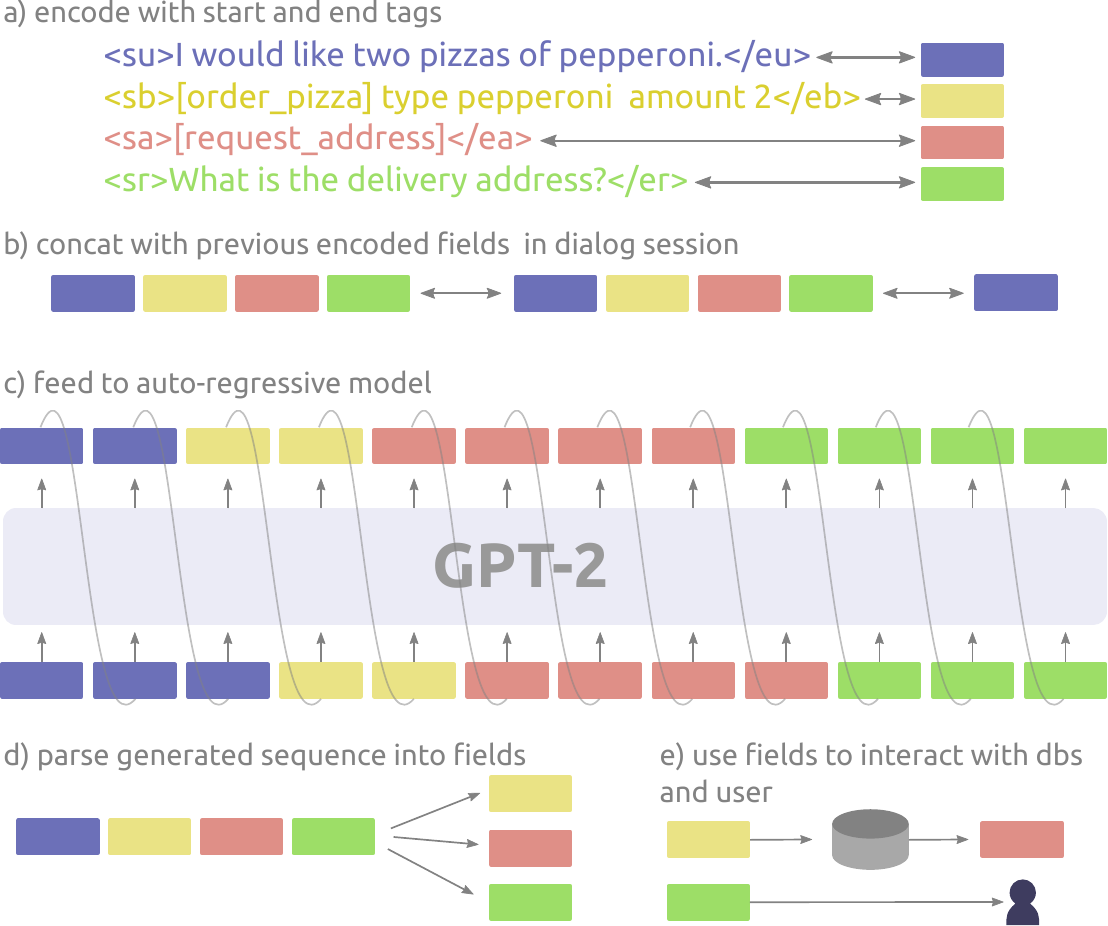}
    \caption{Operational diagram of \ac{tods} based on \acp{lm}. The task is encoded as a sequence-to-sequence problem to later be decoded and parsed for system communication.}
    \label{fig:ubar}
\end{figure}

\ac{tods} are conversational agents designed for solving business-related tasks by interacting with customers through a conversation in a natural language interface. Its task is to solve sub-problems related to business attendance --- identify intentions, and entities, decide which action to take, and generate a system response. As a natural language process, \ac{tods} could benefit significantly from the general language capabilities of \acp{lm} \cite{fellows2021task}. The academy has explored encoding \ac{tods} as a sequence-to-sequence problem \cite{lin-etal-2020-mintl,hosseini2020simple,yang2020ubar} to improve significantly the effectiveness of agents in unrestricted domains. Figure \ref{fig:ubar} illustrates an example of how this process occurs. The general task is composed of four steps; first, it receives a user utterance (denoted by su tag); then it identifies intents and entities in the utterance; the belief state (denoted by sb tag); then it decides which action to take (sa tag); finally it generates a response for this action to return to the user.

In Figure \ref{fig:ubar}, the interactions occur in the following manner: a) we first represent every step in textual symbols, and each field is marked by the start and end of tags; (b) we concatenate the current utterance with the fields from previous turns in the dialog; (c) we feed the whole dialog history to the auto-regressive model to generate the following steps (belief state, action, and response). (d) the generated fields are parsed, then used for querying information; later we return the response to the user as shown in (e).

The \ac{tods} task encoded as sequence-to-sequence presents an unusual structure that is not present in the text used for pre-training language models. The encoded sentence presents a regular grammar, for the meta-structure encoded by unique tag tokens. Utterances and responses have a conversation distribution, so it is not the same distribution of words of a general text domain \cite{zhang-etal-2020-dialogpt}. We have a classification task, where the model aims to infer user intent given his utterance. Then, a \ac{ner} and parsing task, where the model has to extract and format slot-values, and it has a policy that is trying to infer that will shape the response \cite{yang2020ubar, hosseini2020simple}.

General language modeling is well suited for many \ac{nlp} tasks. However, it still requires fine-tuning on large datasets for tasks with complex behavior, like textual distributions that diverge a lot from source training \cite{Liu2021GPTUT, Ouyang2022TrainingLM, Nakano2021WebGPTBQ}, or domain-specific applications \cite{wu-etal-2020-tod, zhang-etal-2020-dialogpt, thoppilan2022lamda}. For \ac{tods}, given the sequential nature of decisions, a high error is not permissible, as it propagates to the following interactions in a dialog, and many examples in the downstream task are needed for proper behavior of conversational agents \cite{fellows2021task}.

Although we could build conversational datasets for our \ac{tods} application, this specialized data is expensive \cite{zhang2020recent}, and small businesses could not afford to build massive datasets. Facing the problem of adaptation with low-resource in the downstream task, we should learn a proper bias to allow the rapid adaptation of our learning model to the drift in downstream data, so the final model should require only a fraction of the usual needs.

Inspired by psycho-pedagogical and optimization literature, a curriculum approach presents examples to a learner that gradually increases in complexity or difficulty \cite{bengio2009curriculum}. This means the learner could focus on acquiring the basic skills in the initial steps that help to learn later posed examples. By the same principle, we could guide the learner through this learning process by presenting not only examples, but new tasks with increasing complexity that are somehow related \cite{green2005integrating}.

In this study, we propose to have intermediate tasks in that we can more easily acquire training data, and this data simulates one or more properties present in the final task. By having another pre-training step to acquire intermediate skills, the model would require fewer fine-tuning steps for the final task, being less prone to over-fit, and acquiring a better generalization. In our approach, we present a transfer learning ``curriculum'', where we gradually adapt our \ac{lm} by posing it to intermediate pre-training tasks. In initial tasks, we have an infinite number of examples; as the complexity of the task increases, the available data decreases exponentially. However, our initial tasks adjust the bias for the \ac{lm} such that it can learn more easily those more complex examples.

We use forum structure to simulate the regular grammar and the word distribution in utterances and responses, and the topic being discussed to simulate the intent classification task. This approach significantly improves \texttt{BLEU}, \texttt{SUCCESS}, and \texttt{INFORM}, compared to other pre-training approaches for \ac{tods}. In this article, we contribute with the following:
\begin{itemize}
    \item We propose a new curriculum learning method where the complexity of learning varies at a sentence task level instead of an instance level. This approach allows for gradually fine-tuning scaled \acp{lm} to small datasets without over-fitting.
    \item As a medium for intermediate training steps, we propose a method for constructing pseudo-supervised data in the context of conversational agents that encompass the main problems in the target task, forcing the \ac{lm} to meta-learn behaviors that help to solve the final target task.
\end{itemize}

The remaining of this article is organized as follows: Section \ref{sec:ref} presents background studies. Section \ref{sec:curriculumtl} formalizes our Curricular Transfer Learning proposal and presents our approach to solving the \ac{tods} task. Section \ref{sec:experiments} presents our conducted experimental evaluation. Section \ref{sec:discussion} discusses the results and Section \ref{sec:conclusion} draws conclusions from our study.

\section{Background} 
\label{sec:ref}

Our research work is heavily inspired by two key studies addresing curriculum learning and transfer learning. We review studies applied to \ac{tods}, but also in general machine learning literature.

\textbf{Curriculum Learning.} Curriculum learning was initially proposed for next word prediction \cite{bengio2009curriculum}, and later extended for many \ac{nlp} tasks \cite{soviany2021curriculum}. In the majority of the studies the curriculum is applied in an instance level \cite{liu-etal-2021-scheduled, kim-etal-2021-document, dai-etal-2021-preview, zhu-etal-2021-combining-curriculum-learning}. Other set of studies, the task is adjusted with a gradual change in the task, where the complexity increases based on length \cite{Foglino2019AnOF} or mode\cite{Liu2020TaskLevelCL}.

In the context of \ac{tods}, some studies apply a curriculum technique in \ac{rl} context, e.g., sort dialogue instance based on number os slots to track \cite{saito-2018-curriculum}, dialog reward information as a measure of complexity \cite{Zhu2021ColdstartedCL, liu-etal-2021-scheduled}, or dialog metrics are used to sort instances in the curriculum \cite{zhao-etal-2022-versatile, dai-etal-2021-preview}.

\textbf{Transfer Learning.} Literature of end-to-end \ac{tods} recently presented studies focusing on rapid adaptation or learning with low resources. Most investigations focus on heuristics for rapid adaptation to data, like variational inference \cite{zhang-etal-2020-probabilistic}, optimizing editing distance \cite{lin-etal-2020-mintl}, multi-tasking \cite{kulhanek-etal-2021-augpt, Su2021}, or denoising \cite{sun-etal-2022-bort}. 

For the conversational literature, there is a set of studies that explore exclusive ``conversation-biased'' data for pre-training language models. Some works explore self-supervision on raw text from Reddit \cite{wu-etal-2020-tod, zhang-etal-2020-dialogpt}, while other present the text as a simple question-answer pair \cite{adiwardana2020humanlike, thoppilan2022lamda}.

We summarize our contribution as follows: In the curriculum learning literature for \ac{nlp}, no study employs a curriculum at a dataset level. For the transfer learning literature, some studies apply pre-training on previous general conversation-oriented data, but do not exploit massive pseudo-supervised data as one of the steps of pre-training. 

Our solution aims to reduce the requirements for annotated data in training sequence-to-sequence tasks. We manipulate the existing structure in web-data to create intermediate tasks that simulate the same sequence structure present in the target task, which we call ``data hacking''. This approach helps the language model to acquire a better bias for the final task. We create data that resembles a simplified version of the target grammar we want our model to acquire, so it can learn basic properties of the grammar and later generalize to more complex instances. In the next section, we present our proposal for complex grammar acquisition by grammar decomposition and instantiate the solution for the problem of \ac{tods}.

\section{Grammar Acquisition with Curricular Transfer Learning}
\label{sec:curriculumtl}

The classical transfer learning framework, using sequential induction, allows us to use out-of-distribution data from a source task to leverage the knowledge in this data for another correlated target task. By learning the proper bias for the problem to be modeled, the learner model can rapidly adapt to the target hypothesis, demanding fewer examples to acquire a good generalization point in the target domain. In the context of \ac{nlp}, transfer learning is achieved by pre-training an \ac{lm}, with self-supervision, in texts sampled from indexed pages on the web \cite{Devlin2019BERTPO, radford2019language, raffel2020exploring}, generally with diverse contents.

Although evidence shows that language modeling acquires high generalization capabilities for the out-of-distribution data \cite{hendrycks2020pretrained}, some specialized and complex tasks that diverge from the general distribution of random web text are less benefited \cite{mielke-etal-2019-kind, cotterell-etal-2018-languages, zellers-etal-2019-hellaswag, deletang2022neural, thoppilan2022lamda}. In low-resource contexts, those pre-trained models tend to overfit or underfit the target task, failing to arrive at a desirable optimum and presenting poor generalization.

In the case of \ac{tods}, the sequence we want to model has a conversation-biased distribution of words, many unique marking tokens for the classification, named entity recognition, and parsing. This drift between pre-training and fine-tuning makes it difficult for our agent to learn and acquire optimal effectiveness \cite{zhang-etal-2020-dialogpt}. Exploring other pre-training data that resemble the same properties in the \ac{tods} task could significantly improve the learner.

We observe the problem of training a model from two perspectives: 1) the learning perspective, where we aim to obtain the model to acquire the pattern recognition skills; 2) the optimization aspect, where we expect that the model minimizes some loss function. As posed by \citep{bengio2009curriculum}, a curriculum approach has a beautiful interpretation and practical application for both perspectives. In our approach, we teach a model in the same way humans learn by gradually increasing the task's difficulty, so it first acquires the basic skills to thrive in more complex scenarios. Also, relying on this curriculum approach, we might minimize a less noisy version of the original problem to arrive at the global optima, a continuation method.

For human language acquisition, the generative grammar theory assumes that a learner has an innate universal grammar that restricts what kind of grammar a learner could acquire \cite{white2003second}. The process starts by recognizing simple structures in a grammar; instead of memorizing, the learner identifies syntactic structures they encounter and evaluates the feedback in an environment to determine precisely the grammar being used \cite{guasti2017language}.

Whereas most of a child's primary language is obtained culturally, without directed control, for second language acquisition the process is almost entirely performed in a controlled environment. In the initial stages, the learner recognizes simple grammatical structures; which is crucial for complete grammar acquisition, as many possible candidate grammars are filtered in this search step \cite{komarova2001evolutionary}. By exploiting this universal grammar, the learner could quickly acquire the grammar of the language it is inserted.

Several approaches in \ac{ai} explore the advantages of posing easier instances of the problem or addressing sequentially surrogate objectives to achieve more complex goals. In \ac{rl}, robotic hand manipulation is an example. Instead of directly training the robot to put a red box over a blue box, it is easier first to teach the arm to recognize which color to pick, then how to hold the box correctly, and later how to place it over the blue box \cite{manela2022curriculum}. This decomposition of complexity allows agents to learn the final task faster than directly attempting to perform the final task.

In the context of sequence-to-sequence modeling, we can view the problem as a grammar acquisition process, starting from basic construction to then acquiring more complex ones. In our developed solution, we first present the learner with basic grammar with a more simple composition, to later extend for more complex elements \cite{davies1980language, guasti2017language}.



We present a formal definition for the problem of grammar acquisition. We extend the model of \citep{pan2009survey} for the multi-sequential case which is an instance of the curriculum learning model.

First, given a domain $\mathcal{D} = \{\mathcal{X}, P(X)\}$ consisting in a feature space $\mathcal{X}$ and a probability distribution $P(X)$, where $X = \{x_1,x_2,...,x_n\} \in \mathcal{X}$. Consider a task $\mathcal{T} = \{\mathcal{Y},f(\cdot)\}$, where $\mathcal{Y}$ is the label space, and $f(\cdot)$ is the objective predictive function, which is not observed but could be learned from the data, where $f:X\rightarrow\mathcal{Y}$. 

We define transfer learning as, given a source domain $\mathcal{D}_S$, a source task $\mathcal{T}_S$, and target domain $\mathcal{D}_T$ with a target task $\mathcal{T}_T$. Transfer learning aims to help improve the learning of the target predictive function $f(\cdot)_T$ in $\mathcal{D}_T$ using the knowledge in $\mathcal{D}_S$ and $\mathcal{T}_S$, where $\mathcal{D}_S \neq \mathcal{D}_T$ or $\mathcal{T}_S \neq \mathcal{T}_T$.

Although most studies in \ac{nlp} literature apply a single step of sequential transfer learning, we propose many steps of transference in the \ac{ctl}. It is a curriculum of transference, \textit{e.g.}, $\boldsymbol{\mathcal{T}} = \{ \mathcal{T}_{S_0}, \mathcal{T}_{S_1},...,\mathcal{T}_{S_n}, \mathcal{T}_T\}$, where we have $n$ pre-training source tasks $\mathcal{T}_S$. By this approach, the $\mathcal{T}_{S_k}$ source task helps optimizing for
$\mathcal{T}_{S_{k+1}}$ task, acquiring a better generalization for the following task and, consequently, the general curriculum as an accumulation of improvements.

Similar to a curriculum learning approach, consider the instances from $\mathcal{T}_{S_k}$ as smoothed objectives of $\mathcal{T}_{T}$. The cost function $C_{S_0}$ is easier to optimize as it does not compass the entire \ac{tods} objective, and we have more available instances in this set. After optimizing for this objective, we gradually increase the task complexity by passing tasks with more linguistic features to extract from $C_{S_0}$ to $C_{S_n}$; and finally $C_{T}$,  fewer instances are presented for more complex sets. 

In the language modeling problem, the cost function we aim to minimize is the cross-entropy for the next token. When the $\lambda$ family is composed of languages with increasingly complex grammar for a cost function $C_0$ that resembles some other target cost function $C_1$ for some family $C_\lambda(\theta)$ of cost functions, the $\lambda$ factor varies between instances according to each data set.

Now, suppose a serialized decoding is learned to a specific pattern. In that case, for instance, the regular grammar of starting and ending fields, adapting to a new grammar that shares the same pattern, will be more straightforward. So the probability of the token to come after \texttt{<eos\_u>} is previously learned in a more manageable task, so the optimization in this phase should only focus on learning the classification task.

Considering that our final task is composed of formal grammar. Suppose another data source could simulate some derivation nodes in the target grammar without loss in the global structure, e.g., the first production rule. In that case, we pose this simpler version of the grammar as a step in the pre-training curriculum, and recursively, append more simplified grammars that simulate previous ones.

We formally call it a curricular transfer learning if, given an ordered set of sequence-to-sequence tasks $\boldsymbol{\mathcal{T}} = \{ \mathcal{T}_S, \mathcal{T}_{I_1},...,\mathcal{T}_{I_n}, \mathcal{T}_T\}$, some complexity ordering $<_\mathcal{C}$, and some grammatical similarity $\sim_G$. Every task in $\boldsymbol{\mathcal{T}}$ respects the order

\begin{equation}
    \mathcal{T}_a <_\mathcal{C} \mathcal{T}_b \hspace{3em} \forall a,b : a<b 
\end{equation}
where $a$ and $b$ are indexes for tasks in $\boldsymbol{\mathcal{T}}$ and
\begin{equation}
    \mathcal{T}_a \sim_G \mathcal{T}_b \hspace{3em} \forall a,b .
\end{equation}

The main appeal of this method is that more accessible instances of the problem have a derisive cost to obtain, whereas final instances have a considerable cost. The artificially created intermediate tasks should help the \ac{lm} to meta-learn the final task, like learning unique tokens to classify, perform a name entity recognition, parsing, or generate a response. We describe the method for complex grammar acquisition in the following manner:

\begin{enumerate}
    \item Describe the desired grammar formally to be learned, e.g. the generation rule for the tokens, listing the expected types of text distribution and regular structure.
    
    \item Find text and structures that can be morphed to simulate this behavior. For the same kind of word distribution in the environment (formal, conversational, etc.), some \ac{nlp} tasks are present in the problem, like classification or NER, etc.
    
    \item Recursively proposes (1) that address the subsequent grammar as another pre-training step.
\end{enumerate}


\section{Experimental Evaluation}
\label{sec:experiments}

We present how we developed our proposed curricular transfer learning steps for the case of \textit{MultiWoZ}. A general intuition for creating intermediate tasks is to find textual corpora or data sets that we transform into a form that resembles properties from final grammar. We demonstrate how we applied our proposal to improve the optimization and overall effectiveness in MultiWoZ, a widely adopted framework for study and evaluation of \ac{tods}. We present a curriculum decomposition with a meta-structure and a single additional task and we compare it with previous pre-training approaches.

\subsection{Protocol}

For the conversation-oriented curriculum, we consider a diverse corpus in the first stage to meta-learn general language skills; a corpus that simulates the conversational patterns and distributions of words; and finally, the data set for the target task. We use the following text data sets as pre-training steps:

\begin{enumerate}

    \item \textbf{Common Crawl (General Language Modeling)} This dataset was proposed as the unsupervised auto-regressive pre-train for the proposed architecture \cite{radford2019language}. Although we do not train the GPT-2 from scratch in this dataset, we describe it here to illustrate the \ac{ctl} approach. 
    
    \item \textbf{Tripadvisor (Dialog  Modeling)} We built this data set by crawling major cities discussed in Tripadvisor (Paris, Rome, Istanbul, Barcelona, Madrid, Amsterdam, Lisbon, and London), for each thread we encoded the thread creator with user utterance tokens, the city as the belief tokens, and thread replies as system response. We replicate each pseudo-user utterance for each pseudo-system response. The crawler is available in Github\footnote{\url{https://github.com/jadermcs/tripadvisor-dialogues}}.
  
    \item \textbf{Multi-Domain Wizard-of-Oz (Task-Oriented Modeling)} The MultiWoZ is a general domain task-oriented dialogues data set collected in a Wizard-of-Oz setting. It comprises 10000 dialogues in domains like `restaurant,' `train,' and `hotel booking.' We selected the latest official data set for our evaluation, version 2.2.
    
\end{enumerate}

To our \ac{lm} learn how to solve this problem, we need to encode it as a sequence-to-sequence prediction problem. To this end, we use the special \texttt{<sos\_X>}, where $X \in {u,b,a,r}$ for the start and end of the string in utterance, belief, action, and response tokens. The example below represents the tokens dialog from \textit{MultiWoZ}, encoded as a sequence-to-sequence problem:

\begin{plaintext}
    <sos\_u> I am looking for [...] <eos\_u> <sos\_b> [restaurant] pricerange cheap area
    centre <eos\_b> <sos\_a> [inform] choice [request] food <eos\_a> <sos\_r> There are [value\_count] restaurants [...] <eos\_r> <sos\_u> ...
\end{plaintext}

We first formalize the target grammar to enlist known properties of this sequence. We notice that we have a regular grammar for demarking field tokens, in long sequences if our model is unable to predict this simple pattern we might fail to parse the decoded sequence.

As the training example is the concatenation of the entire dialog session, the language of this grammar should be any form in $\mathcal{L}=\{\epsilon, UBAR, UBARUBAR,...\}$, cycling infinitely in the regular pattern.

For each field in our language $\mathcal{L}$, we have a grounded theory for linguistic patterns that are present. Fields $u$ and $r$ follow a dialog act distribution \cite{traum1992conversation, Bunt2006DimensionsID}. We have that string $w$ in $u$ is sampled from $u(w)$ an utterance distribution. The string $w$ in $R$ is sampled from a response distribution of words $r(w)$. \texttt{<se*>} are delimiter tokens for each segment. 

In the $b$ field, we have a classification task, and a \ac{ner} with parsing as a sequence problem so $k \subset K_b; i \subset I_b$. In $a$, our model aims to learn a policy optimization \cite{williams2016dialog}, $k \subset K_a; i \subset I_a$. Table \ref{tab:grammar} illustrates the grammar our model aims to acquire.

\begin{table}[ht]
    \centering
    \caption{Regular grammar for \ac{tods} encoding in a language modeling problem.}
    \label{tab:grammar}
    \begin{tabular}{rl}
        $S$ & $\rightarrow U B A R S$ | $\epsilon$ \\
        $U$ & $\rightarrow$ \texttt{<su>}$w_u \sim u(w_u)$\texttt{<eu>} \\
        $B$ & $\rightarrow$ \texttt{<sb>}$k \subset K_b; i \subset I_b$\texttt{<eb>} \\
        $A$ & $\rightarrow$ \texttt{<sa>}$k \subset K_a; i \subset I_a$\texttt{<ea>} \\
        $R$ & $\rightarrow$ \texttt{<sr>}$w_r \sim r(w_r)$\texttt{<er>} \\
    \end{tabular}
\end{table}

If we simulate nodes $U$, $B$, and $R$ we obtain a candidate grammar for an intermediate task. To obtain this data, we crawled conversations from online forums
where a user has a centered topic discussion and starts a thread by posing a question. Later, users can join the discussion by replying to the thread creator or other messages.

We manipulate these data to resemble our grammar in the following manner: we set the message that creates the thread as an utterance and the topic of this conversation as the classification problem. We put an empty string for action, and for the response, we pick an answer in the thread. We replicate this pattern for every message replying to the original topic creator. Finally, we append the unique tokens to encode it. This process is fully automated by a script using the HTML annotation from scraped data. We present the template below:

\begin{plaintext}
    <sos\_u> i'll be in amsterdam for a week [...] <eos\_u> <sos\_b> amsterdam <eos\_b> <sos\_a> <eos\_a> <sos\_r> take your pick [...] <eos\_r> 
\end{plaintext}

Table \ref{tab:pseudogrammar} illustrates the formal version of our fabricated grammar that simulates an intermediate transfer step.

\begin{table}[ht]
    \centering
    \caption{Regular grammar for simulating the distribution of the target grammar.}
    \label{tab:pseudogrammar}
    \begin{tabular}{rl}
        $S$ & $\rightarrow U B A R S$ | $\epsilon$ \\
        $U$ & $\rightarrow$ \texttt{<su>}$w_u \sim u(w_u)$\texttt{<eu>} \\
        $B$ & $\rightarrow$ \texttt{<sb>}$\boldsymbol{k \subset K_c}\textcolor{gray}{; i \subset I_b}$\texttt{<eb>} \\
        $A$ & $\rightarrow$ \texttt{<sa>}$\textcolor{gray}{k \subset K_a; i \subset I_a}$\texttt{<ea>} \\
        $R$ & $\rightarrow$ \texttt{<sr>}$w_r \sim r(w_r)$\texttt{<er>} \\
    \end{tabular}
\end{table}

In gray, we show the removed elements, and in bold pseudo-elements. The final step is to randomly concatenate the above template with random messages from different discussion topics, so it can generalize the regular pattern.

To better evaluate our transference curriculum, we proposed four different training settings. The first is the used in \citep{yang2020ubar}. The second approach is to pre-train it in a conversational distribution without any special encoding. Finally, the last curriculum encodes all the grammar in Table \ref{tab:pseudogrammar}. We specify the curriculum details in the following:

\begin{enumerate}
    \item The first curriculum named ``gpt-*/multiwoz,'' uses pre-trained weights from a GPT-2 from \textit{HuggingFace}.
   
    \item The second curriculum named ``gpt-*/noencode/multiwoz,'' starts with GPT-2 weights, then we train it on TripAdvisor data, with the random ordering of the texts and no special encoding.
    
    \item The third curriculum named ``gpt-*/encode/multiwoz,'' is the fully encoded task modeling utterance, response, and classification task for the pre-training.
    
\end{enumerate}

For all curriculums that include \textit{TripAdvisor}, we pre-process the text using the same script for \textit{MultiWoZ}. We train this architecture in the same strategy as \cite{yang2020ubar}, where all the dialog session is presented to the model in a single sequence. The training sequence should contain at most 256 tokens, sessions with more than this maximum length we split into the following sequence. We train the models with early stopping with a plateau of 5 steps.

For the inference stage, we decode tokens with a sampling strategy normalizing with $\tau = 0.7$ until we reach an end-of-response token. We give belief state and action from the oracle and just decode the response with the slot-values tokens.

We evaluate the proposed curriculum under two perspectives: (i) the running loss for each curriculum variation; and (ii) the standard metrics for the MultiWoZ data sets with oracle values for database and action, computed by \cite{nekvinda-dusek-2021-shades} library.

In the first analysis, we investigate how the initialization proposed by each curriculum helps in the optimization process for minimizing the empirical loss for the target grammar. The second analysis measure, given each curriculum, how this improved minimization translates into better agent performance for the \ac{tods} task. In this step, we compute using \texttt{BLUE}, \texttt{INFORM}, and \texttt{SUCCESS}.

\subsection{Results}
\label{sec:results}

Our evaluation consists of two experiments, curriculum convergence, and optimality. For convergence, we vary the model size and evaluate the scalability and time required to stop training in the curriculum. 

\begin{figure}[ht]
    \centering
    \includegraphics[width=\columnwidth]{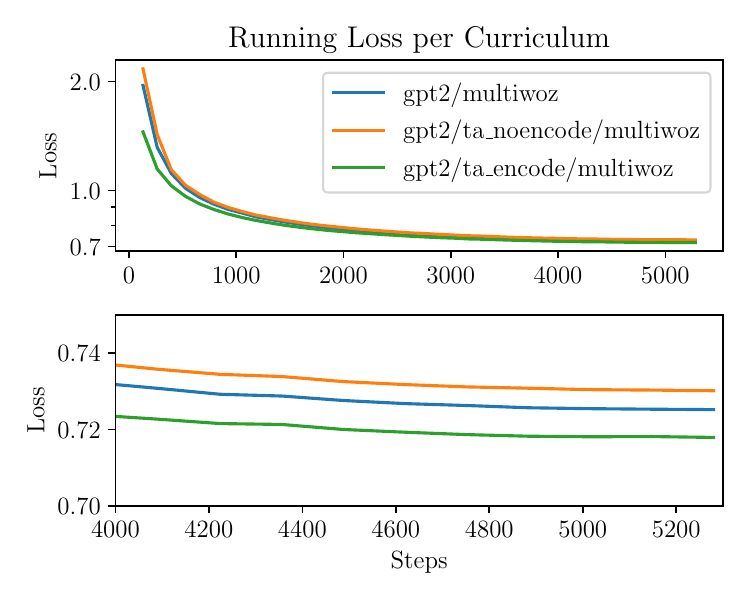}
    \caption{Loss during training on validation split of MultiWoZ. Each line represents one curriculum configuration. The bottom figure is a zoom in the final steps of early stopping.}
    \label{fig:running_loss}
\end{figure}

Figure \ref{fig:running_loss} presents the loss for the validation split over time. We observe that the curriculum with the pseudo-tods task has a significantly better starting point for the MultiWoZ task and converges to a lower loss than the other curriculums. For the small architecture, it converges faster.

Table \ref{tab:metrics} presents the \ac{tods} metrics computed for the test split. We observe that the pseudo-tods curriculum performs better than other curricula. The \texttt{COMBINED} score is given by $(\texttt{INFORM}+\texttt{SUCCESS})*0.5+\texttt{BLEU}$.

\begin{table}[ht]
    \centering
    \caption{Standard metrics for the test split in MultiWoZ. Each row contain a curriculum that is evaluate in different model sizes.}
    \resizebox{\columnwidth}{!}{
    \begin{tabular}{|r|c|c|c|c|c|}\hline
        Curriculum & Model Size & \texttt{BLEU} & \texttt{INFORM} & \texttt{SUCCESS} & \texttt{COMBINED} \\\hline
        
        No & & \textbf{31.3}& 93.4 & \textbf{90.4} & 123.2 \\
        No-encode & Small & \textbf{31.3} & 93.3 & \textbf{90.4} & 123.2 \\
        CTL (ours) & & \textbf{31.3}& \textbf{93.6} & \textbf{90.4} & \textbf{123.3}\\ \hline

        No & & 31.7& 93.2 & 90.3 & 123.5 \\
        No-encode & Medium & \textbf{31.8} & 93.3 & 90.2 & 123.6 \\
        CTL (ours) & & 31.7& \textbf{93.5} & \textbf{90.5} & \textbf{123.7}\\\hline

        No & & 31.5& \textbf{93.3} & 89.6 & 123.0 \\
        No-encode & Large & 31.0& 93.0 & 89.3& 122.2\\
        CTL (ours) & & \textbf{31.6} & \textbf{93.3} & \textbf{89.9} & \textbf{123.2} \\\hline
    \end{tabular}}
    \label{tab:metrics}
\end{table}

We found that the faster adaptation and overall loss observed during training also translated to the high-level metrics; and the larger the model, the more consistent the gain.

\section{Discussion}
\label{sec:discussion}

Differing from traditional language modeling, where the task is to predict sequences of words given a context, a task-oriented dialog presents a structured sequence prediction problem with three \ac{nlp} sub-tasks: (i) a classification task to recognize intents; (ii) a \ac{ner} to recognize  entities; and (iii) a natural language generation task to predict the adequate response for a given utterance and system actions. \acp{lm} are general multi-task learners for \ac{nlp}, which overfit on specific contexts with a nonstandard distribution of tokens.

In this research, we defended that having a general language modeling task as a starting point is not ideal for the problem of \ac{tods}. As our problem has a strong grammar dependency, it is, the generated text should be parseable, and the model should always predict the same tokens even if the dialog session has a hundred interactions. This requires constructing gradual datasets to better initialize the \ac{lm} and guarantee it does not degenerate in predicting the grammar for long sequences. 

Our proposed solution, \ac{ctl}, a sequence of transfer learning steps, helps to balance the trade-off between the right bias in the data and annotation costs. Our proposal shows how this approach is not only a multi-step sequential transfer learning, and it can be viewed as a form of curriculum learning when considering the overall ordering within all tasks. We optimized order through the same data-generating process in the original curriculum learning approach. Our \ac{ctl} approach allows us to use out-of-distribution data, which is central to learning with scalability and generality for modern \ac{nlp} systems.

In our experiments we could verify that our proposed pre-training approach signicantly improve the initialization for training on MultiWoZ data, resulting in an overall improvement in standard \ac{tods} metrics when compared to previous approaches.

Our approach explored the \ac{ctl} on a narrow case study, given that sequence-to-sequence task encoding is a relatively new proposal \cite{hosseini2020simple}. However, prompt-based methods are a recent trending topic in \ac{nlp}. These methods allow a pre-trained \ac{lm} to perform out-of-the-box several tasks they were not explicitly trained for (also called zero-shot learning). Although those models perform pretty well in classic \ac{nlp} tasks, those with a complex morphological structure still need additional learning. We could extend those models with \ac{ctl} to other complex applications by creating pseudo-labeled datasets in a super-scale.

\section{Conclusion}
\label{sec:conclusion}


In this research, we developed a solution to train \acp{lm} for complex grammar acquisition. Our proposal was accomplished by training the model in intermediate datasets following the grammar simplification method. This allows external data sources to encode as an intermediate sequence task to improve the general optima. Unlike earlier research attempts to use additional data depending on annotated data, which is expensive and not feasible for many business models, our approach used unsupervised data to structure pseudo-supervised data and model sharing to significantly reduce the costs of maintaining this kind of system.


\bibliography{anthology,custom}
\bibliographystyle{acl_natbib}

\end{document}